\documentclass{article}
\usepackage{spconf,amsmath,amssymb,graphicx,booktabs,multirow}
\usepackage{stfloats}
\usepackage{float}
\usepackage{url}
\usepackage{flafter,placeins}
\usepackage{xcolor}
\usepackage{colortbl}
\usepackage{bm}
\newcommand{\p}{\ensuremath{\mathbf{P}}}

\makeatletter
\ninept
\makeatother

\title{Complexity-Calibrated Reconstruction Discrepancy for\\ No-Reference Point Cloud Quality Assessment}
\name{Haocheng Wu, Xiangjie Sui$^{\dagger}$\thanks{$^{\dagger}$Corresponding author (email: xjsui@cityu.edu.mo).}, and Xin Sun}
\address{City University of Macau, Macau SAR, China}

\begin{document}
\maketitle

\begin{abstract}
The discrepancy induced by surface reconstruction in a distorted point cloud is jointly influenced by its intrinsic structural complexity and signal degradation. This observation motivates a reference-free point cloud quality assessment (PCQA) strategy  that estimates and compensates for the complexity-induced component. We implement this strategy through successive reconstruction. Specifically, reconstructing the distorted point cloud produces an input-to-reconstruction discrepancy that combines the effects of signal degradation and content complexity. Reapplying the reconstruction operator with stronger smoothing produces a discrepancy between successive reconstructions, which serves as a proxy for the complexity-induced component. This proxy is then used to calibrate the input-to-reconstruction discrepancy. We apply the same strategy to 3D geometry patches and their aligned 2D regions, enabling local comparison of the resulting discrepancies. The calibrated geometric and projected-view discrepancies are then fused into a final degradation score. For both branches, the discrepancies are measured using frozen pretrained perceptual representations. Our method does not require PCQA-specific training or fine-tuning. Experiments on three public databases show that the proposed method consistently outperforms most generic baselines and achieves performance comparable to several full-reference metrics.The code is available at \url{https://github.com/HC-W-whc/CCRD-NR-PCQA.git}

\end{abstract}


\begin{figure}[!t]
\centering
\includegraphics[width=1\columnwidth]{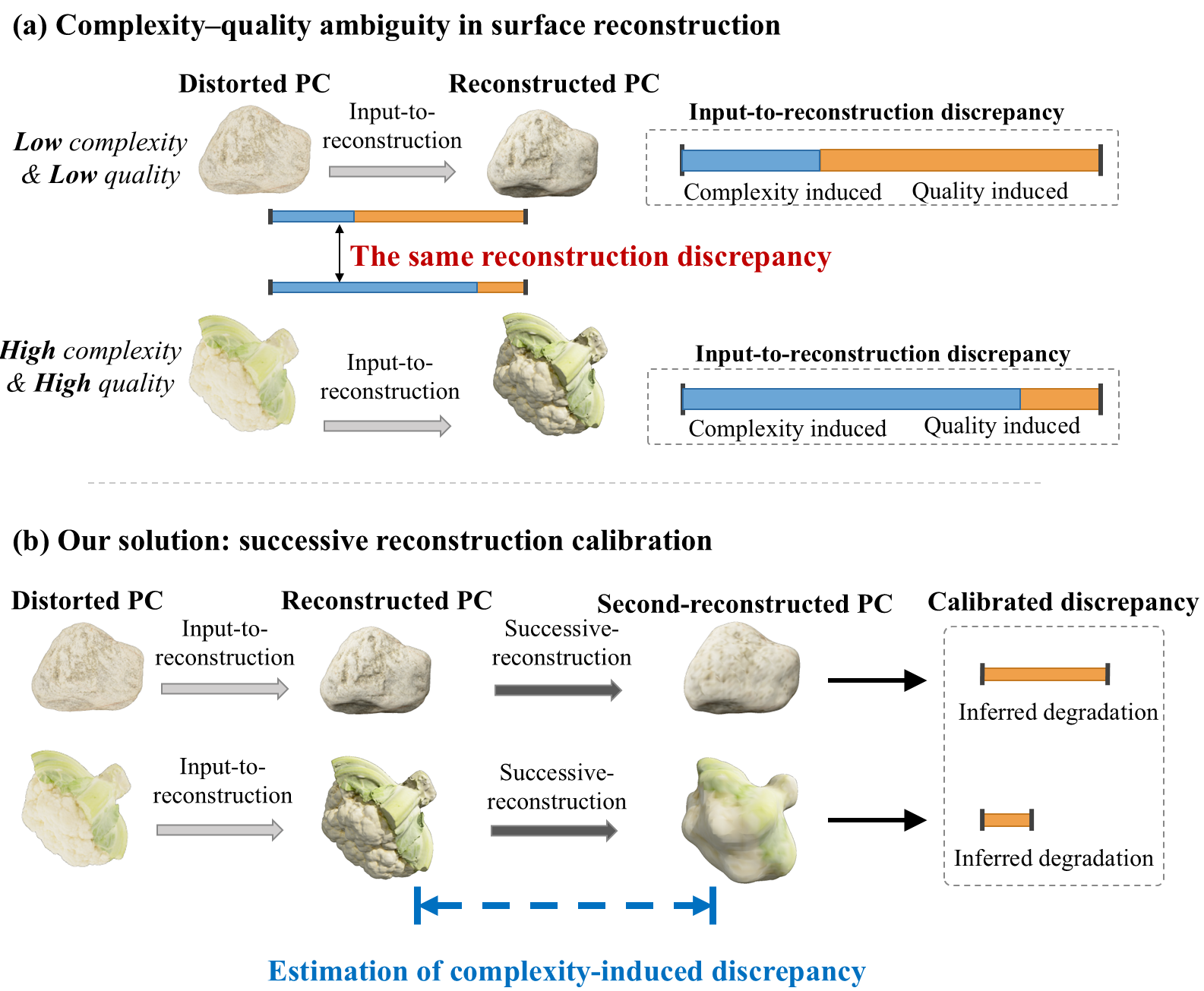}
\caption{Motivation for successive-reconstruction calibration in reference-free point cloud quality assessment. (a) The input-to-reconstruction discrepancy mixes intrinsic structural complexity with signal degradation, leading to ambiguity in quality inference. (b) A second reconstruction with stronger smoothing is used as a proxy for the complexity-induced component and enables calibration of the initial discrepancy.}
\label{fig:intro}
\end{figure}

\section{Introduction}
\label{sec:intro}

Point clouds are irregular 3D signals whose geometry and appearance may be degraded during acquisition, compression, transmission, and rendering~\cite{8571288}. Quantifying the perceptual impact of these impairments is essential for developing and optimizing point cloud processing systems. Subjective assessment provides reliable perceptual scores~\cite{9238424,10.3389/frsip.2024.1420060}, but is impractical for automatic quality monitoring. Objective quality metrics provide a practical alternative. Full-reference (FR) quality metrics~\cite{9106005,9123147} compare a distorted point cloud with its aligned pristine counterpart. However, such a reference is often unavailable in practical applications, motivating no-reference point cloud quality assessment (NR-PCQA)~\cite{10.1145/3550274,10381826}.

Most NR-PCQA methods address the missing-reference problem by learning a direct mapping from statistical or deep representations to subjective scores~\cite{9810024,10542438,10814096}.
Projection-based methods further use 2D views to model the perceptual effects of point cloud distortions~\cite{9238424,9496633,10.1145/3643817,Seo_2026_CVPR}. Other methods combine geometric and projected-view information through multimodal architectures~\cite{ijcai2023p195}. Probabilistic models account for the variability of subjective judgments~\cite{ijcai2025p462}. Another line of work constructs a pseudo-reference from the distorted input~\cite{10167694} or restores visual features in projected 2D views before feeding them to a supervised predictor~\cite{MU2024122953}. These studies demonstrate that recovery-related information provides useful quality cues.  However, the discrepancy induced by surface reconstruction in a distorted point cloud is jointly influenced by its intrinsic structural complexity and signal degradation. As illustrated in Fig.~\ref{fig:intro}~(a), point clouds with different combinations of complexity and quality can produce similar input-to-reconstruction discrepancies, making direct quality inference ambiguous.

We address this ambiguity through successive reconstruction, as summarized
in Fig.~\ref{fig:intro}~(b). By viewing surface reconstruction as a regularized transformation~\cite{Huang_2023_CVPR}, we reapply the reconstruction operator with stronger smoothing to expose the sensitivity of distorted point cloud to additional regularization. The resulting discrepancy between successive reconstructions can serve as a proxy for estimating the complexity-induced component in quality inference. To capture complementary geometric and appearance cues, we apply this calibration to aligned local 3D patches and their corresponding 2D patches, allowing their discrepancies to be compared at matched locations. We measure these discrepancies using frozen pretrained perceptual representations~\cite{10.1007/978-3-031-20086-1_35,Zhang_2023_CVPR} and fuse the calibrated geometric and projected-view terms into a degradation score.

The main contributions of this work are:
\begin{itemize}
    \item We identify the confounding effect of content complexity in reconstruction-induced discrepancies and introduce successive reconstruction to estimate this component for reference-free quality inference of point cloud.
    \item We extend complexity calibration from geometric reconstruction discrepancies to matched 2D-view quality changes, providing a unified treatment of reconstruction-induced variation across 3D and 2D observations.
    \item We develop a training-free PCQA framework based on frozen pretrained perceptual representations, eliminating PCQA-specific training and fine-tuning, and validate it on three public databases.
\end{itemize}



\section{Method}
\label{sec:method}
\subsection{Overview}
\begin{figure}[!t]
\centering
\includegraphics[width=\columnwidth]{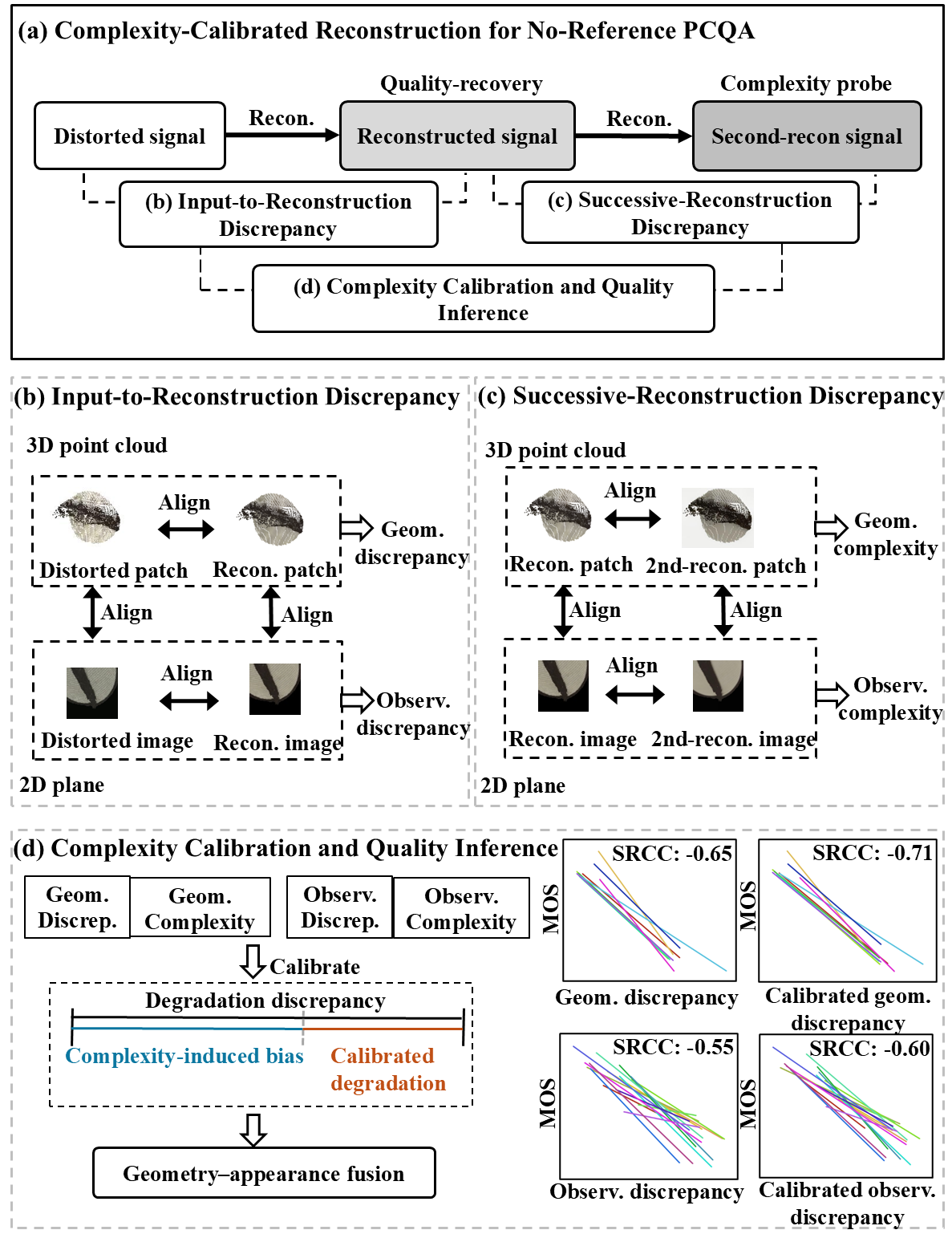}
\vspace{-2.5em}
\caption{Overview of the proposed complexity-calibrated reconstruction framework for no-reference point cloud quality assessment.}
\label{fig:fw}
\vspace{0.8em}
\end{figure}

As illustrated in Fig.~\ref{fig:fw}, our framework uses successive
surface reconstruction to estimate and calibrate reconstruction-induced
discrepancies. Given a distorted point cloud $\p$, we construct two successive reconstructions:
\begin{equation}
\begin{aligned}
&\p'=\mathbf{\Phi}(\p;\alpha_1), \\
&\p''=\mathbf{\Phi}(\p';\alpha_2),  
\end{aligned}
\end{equation}
where $\mathbf{\Phi}$ denotes the surface-reconstruction operator and $\alpha_2>\alpha_1$ controls the reconstruction strength. 
We define the input-to-reconstruction discrepancy and the
successive-reconstruction discrepancy as
\begin{equation}
\begin{aligned}
&\mathbf{\Delta}_{i2r}=\mathcal{D}(\p,\p'), \\
&\mathbf{\Delta}_{r2r}=\mathcal{D}(\p',\p''),
\end{aligned}
\end{equation}
where $\mathcal{D}$ denotes the branch-specific discrepancy measure.
The calibrated discrepancy is then computed as
\begin{equation}
\label{eq:calibration}
\hat{\mathbf{\Delta}}=\mathbf{\Delta}_{i2r}-\lambda\mathbf{\Delta}_{r2r}
\end{equation}
where $\lambda$ controls the strength of complexity calibration. The same formulation is applied to the geometric and 2D appearance
representations. Their calibrated discrepancies $\hat{\mathbf{\Delta}}^{3D}, \hat{\mathbf{\Delta}}^{2D}$ are finally fused into a degradation score. The specific discrepancy measures for the
two representations are introduced in the following subsections.  When evaluated on a local patch or projected region, the same notation is indexed by the local anchor and view.


\subsection{Successive Surface Reconstruction}
We use a pretrained reconstruction model~\cite{Huang_2023_CVPR} to instantiate the operator $\mathbf{\Phi}$, without additional training or fine-tuning on PCQA data. The discrepancy $\mathbf{\Delta}_{i2r}$ reflects both signal degradation and the structural
complexity of the underlying content. In contrast, the discrepancy $\mathbf{\Delta}_{r2r}$ measures the additional change induced by stronger
regularization. Since this change is induced after the initial
reconstruction, it provides a content-dependent cue for estimating the
complexity-induced component of the initial discrepancy. Note that neither $\p'$ nor $\p''$ is treated as a pristine reference. Instead, they form a sequence of observations under increasing regularization.

\subsection{Local 3D--2D Correspondence}
To obtain local quality cues, we sample $K$ anchors from the distorted point cloud $\p$ using farthest-point sampling. Each anchor defines a local geometry patch formed by its $M$ neighboring points. The same anchors are projected onto multiple orthographic views to identify their corresponding 2D appearance regions. This construction preserves the local correspondence between the two representations, allowing their discrepancies to be evaluated at matched locations. We apply the successive-reconstruction procedure to each local 3D patch and its corresponding projected regions. The resulting calibrated geometric and appearance discrepancies are fused at each matched location.

\subsection{Geometric Discrepancy}

For anchor $n$, let $\p_n$, $\p'_n$, and $\p''_n$ denote the corresponding local geometry patches extracted from $\p$, $\p'$, and $\p''$, respectively. We instantiate the discrepancy operator $\mathcal{D}$ with features extracted from a frozen encoder~\cite{10.1007/978-3-031-20086-1_35}.
For the input-to-reconstruction pair $(\p_n,\p'_n)$, let
$\mathbf{z}_{i,n}$ and $\mathbf{z}'_{i,n}$ denote the
LayerNorm-normalized tokens extracted from transformer block $i$. We
measure their directional and magnitude discrepancies as
\begin{equation}
d^{\mathrm{dir}}_{i,n}
=
\frac{\left\|\mathbf{z}_{i,n}-\mathbf{z}'_{i,n}\right\|_{2}}
{\left\|\mathbf{z}_{i,n}\right\|_{2}
+\left\|\mathbf{z}'_{i,n}\right\|_{2}+\epsilon},
\end{equation}

\begin{equation}
d^{\mathrm{mag}}_{i,n}
=
\frac{1}{C}\sum_{c=1}^{C}
\frac{\left|\phi(z_{i,n,c})-\phi(z'_{i,n,c})\right|}
{\max\{\phi(z_{i,n,c}),\phi(z'_{i,n,c})\}+\epsilon},
\end{equation}
where $\phi(x)=\log(1+\exp(x))$ is the softplus function, $C$ is the
feature dimension, $c\in\{1,\ldots,C\}$ indexes the feature dimensions, and $\epsilon$ is a small constant for numerical
stability. The geometric discrepancy at anchor $n$ is obtained by aggregating the two
terms across the selected transformer blocks:
\begin{equation}
\mathbf{\Delta}^{3D}_{i2r,n}(\p_n,\p'_n)
=
\sum_i w_i
\left[
w_{\mathrm{dir}}d^{\mathrm{dir}}_{i,n}
+
w_{\mathrm{mag}}d^{\mathrm{mag}}_{i,n}
\right],
\end{equation}
where $w_i$ denotes the block weight, and
$w_{\mathrm{dir}}$ and $w_{\mathrm{mag}}$ balance the directional and magnitude terms. The same computation is applied to the successive-reconstruction pair
$(\p'_n,\p''_n)$ by replacing
$(\mathbf{z}_{i,n},\mathbf{z}'_{i,n})$ with
$(\mathbf{z}'_{i,n},\mathbf{z}''_{i,n})$, resulting in $\mathbf{\Delta}^{3D}_{r2r,n}$. 


\subsection{Projected-View Discrepancy}
The surface reconstruction operator~\cite{Huang_2023_CVPR}  changes the geometry but does not directly recover the associated color. We  instantiate the same successive-response principle with a two-stage geometry-aware color filtering process. The first stage produces colored 2D projected view $I'$, while the second stage applies stronger smoothing and produces $I''$. Specifically, for a center point $\mathbf{x}$, let $\mathcal{N}(\mathbf{x})$ denote its neighborhood, and let $\mathbf{y}\in\mathcal{N}(\mathbf{x})$ denote a neighboring point.
We compute the filtering weight for stage $t\in\{1,2\}$ as
\begin{equation}
\omega_t(\mathbf{x},\mathbf{y})
=
\exp\left[
-\frac{1}{2}
\left(
\frac{d_{\mathrm{3D}}^2(\mathbf{x},\mathbf{y})}{\sigma_{s,t}^2}
+
\frac{\Delta E_{\mathrm{Lab}}^2(\mathbf{x},\mathbf{y})}{\sigma_{c,t}^2}
+
\frac{\delta_n^2(\mathbf{x},\mathbf{y})}{\sigma_{n,t}^2}
\right)
\right],
\end{equation}
where $d_{\mathrm{3D}}$, $\Delta E_{\mathrm{Lab}}$, and $\delta_n$ denote the spatial, color, and normal differences, respectively. 
$\sigma_s$, $\sigma_c$, and $\sigma_n$ denote the corresponding scale parameters. The filtered color at stage $t$ is then obtained by normalized weighted
averaging:
\begin{equation}
\mathbf{c}^{(t)}(\mathbf{x})
=
\frac{
\sum_{\mathbf{y}\in\mathcal{N}(\mathbf{x})}
\omega_t(\mathbf{x},\mathbf{y})\mathbf{c}(\mathbf{y})
}{
\sum_{\mathbf{y}\in\mathcal{N}(\mathbf{x})}
\omega_t(\mathbf{x},\mathbf{y})
},
\qquad t\in\{1,2\}.
\end{equation}

The second filtering stage applies stronger smoothing by using larger filtering scales than the first stage. As described in Sec.~2.3, each local geometry patch is associated with multiple projected-view regions. Using the common geometry of $\mathbf{P}$, the original color field and the two filtered color fields are projected onto these views, producing $I$, $I'$, and $I''$, respectively. Let $q(\cdot)$ denote the normalized perceptual quality score produced by a frozen pretrained 2D quality model~\cite{Zhang_2023_CVPR}. We use the signed difference between the quality scores of two observations as their projected-view discrepancy. We retain the sign of the quality-score difference because it captures the direction of the perceptual change caused by filtering.  For the $n$-th local geometry patch and its $v$-th projected view, let $N^{\mathrm{fg}}_{n,v}$ denote the number of projected foreground points in the corresponding 2D region. We retain the view if $N^{\mathrm{fg}}_{n,v}\geq N_{\min}$ and define the set of valid views as: $\mathcal{V}_n = \left\{ v\mid N^{\mathrm{fg}}_{n,v}\geq N_{\min} \right\}$. For every $v\in\mathcal{V}_n$, we compute the two projected-view
discrepancies as $\mathbf{\Delta}^{2D}_{i2r,n,v}=q(I'_{n,v})-q(I_{n,v}), \mathbf{\Delta}^{2D}_{r2r,n,v}=q(I'_{n,v})-q(I''_{n,v})$. These view-level discrepancies are calibrated and aggregated in Sec.~2.6. 


\subsection{Complexity Calibration and Quality Fusion}

We apply the same calibration rule (see Eq.~\ref{eq:calibration}) to the geometric and projected-view branches. For the $n$-th local patch, the calibrated geometric discrepancy is
\begin{equation}
\hat{\mathbf{\Delta}}^{3D}_{n}
=
\mathbf{\Delta}^{3D}_{i2r,n}
-
\lambda\mathbf{\Delta}^{3D}_{r2r,n}.
\end{equation}
For the projected-view branch, each valid view is calibrated separately, and the calibrated values are then averaged:
\begin{equation}
\hat{\mathbf{\Delta}}^{2D}_{n}
=
\frac{1}{|\mathcal{V}_n|}
\sum_{v\in\mathcal{V}_n}
\left(
\mathbf{\Delta}^{2D}_{i2r,n,v}
-
\lambda\mathbf{\Delta}^{2D}_{r2r,n,v}
\right).
\end{equation}
The calibrated geometric and projected-view discrepancies are fused to obtain the local degradation response:
\begin{equation}
\hat{\mathbf{\Delta}}_n
=
w_{3D}\hat{\mathbf{\Delta}}^{3D}_{n}
+
w_{2D}r_n\hat{\mathbf{\Delta}}^{2D}_{n},
\end{equation}
where $r_n$ is a reliability factor determined by the number of valid views. If no valid projected view is available, the second term is omitted. Finally, the sample-level degradation is obtained by averaging the local responses:
\begin{equation}
\hat{\mathbf{\Delta}}=\frac{1}{K}\sum_{n=1}^{K}\hat{\mathbf{\Delta}}_n,
\end{equation}
where a larger $\hat{\mathbf{\Delta}}$ value indicates a worse quality.

\section{Experiments}
\label{sec:exp}

\subsection{Experimental setup}

\textbf{Databases.} We evaluate the proposed method on three public PCQA databases. SJTU-PCQA~\cite{9238424} contains $9$ reference point clouds and $378$ distorted samples generated from $7$ distortion types at $6$ severity levels, including compression, color noise, geometry noise, downsampling, and their combinations. WPC2.0~\cite{9490512} contains 16 reference point clouds and $400$ V-PCC samples produced by combining $5$ geometry and $5$ attribute quantization levels, providing controlled geometry--attribute compression distortions. ICIP2020~\cite{9191308} contains $6$ reference point clouds and $90$ samples encoded by G-PCC Octree, G-PCC Trisoup, and V-PCC at $5$ quality levels, primarily covering codec-induced compression artifacts. 

\textbf{Evaluation criteria.}
We use the Spearman rank-order correlation coefficient (SRCC) and the Pearson linear correlation coefficient (PLCC) to measure the agreement between the predicted degradation and the mean opinion scores (MOS). Since a larger distance indicates lower perceptual quality, the predicted distance typically has a negative correlation with MOS. We therefore report $|$SRCC$|$. Before computing PLCC, we apply the standard five-parameter logistic mapping. To fairly compare methods that require training, we use a nine-fold evaluation protocol for each database. In each fold, supervised NR-PCQA methods are trained on the corresponding training split and evaluated on the held-out test split. All methods are evaluated on the same nine test splits, and the reported SRCC and PLCC values are averaged over the nine folds.

\textbf{Implementation details.}
The surface-reconstruction operator, geometric feature extractor, and projected-view quality model are initialized with publicly available pretrained weights and kept frozen throughout the experiments. No PCQA-specific training or fine-tuning is performed. We sample $K=64$ anchors from each distorted point cloud and extract $M=32$ points around each anchor to form a local geometry patch. Each local patch is associated with six orthographic views. A projected view is retained when at least  $N_{min}=32$ points in the corresponding patch is visible. The geometric discrepancy combines the directional and magnitude terms with weights $w_{\mathrm{dir}}=0.2$ and $w_{\mathrm{mag}}=0.8$. The complexity-calibration coefficient is fixed to $\lambda=0.25$, and the geometric and projected-view responses are fused with weights $w_{3D}=0.6$ and $w_{2D}=0.4$. All parameters are fixed across the three databases.


\begin{figure*}[!th]
\centering
\includegraphics[width=0.9\textwidth]{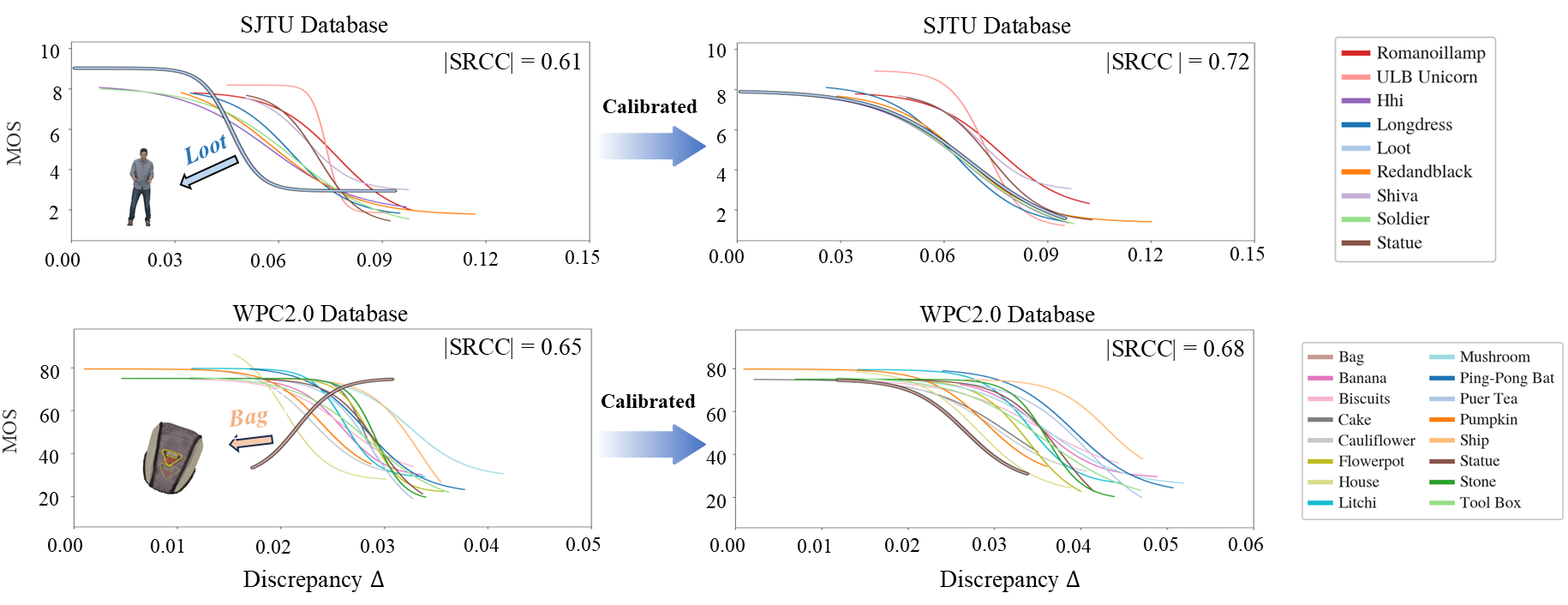}
\caption{Effect of complexity calibration. Curves show the relationship between the reconstruction-induced distance and subjective quality for different source contents on SJTU-PCQA and WPC2.0 databases, before (left) and after (right) calibration.}
\label{fig:content}
\end{figure*}

\begin{table}[!t]
\centering
\caption{Performance comparison of FR PCQA, supervised NR-PCQA and no PCQA-specific training methods. Bold values indicate the best performance within no PCQA-specific training methods.}
\label{tab:cmp}

\resizebox{\columnwidth}{!}{
\begin{tabular}{c l cc cc cc}
\toprule
\multirow{2}{*}{Type}
& \multirow{2}{*}{Methods}
& \multicolumn{2}{c}{SJTU-PCQA}
& \multicolumn{2}{c}{WPC2.0}
& \multicolumn{2}{c}{ICIP2020} \\

\cmidrule(lr){3-4}
\cmidrule(lr){5-6}
\cmidrule(lr){7-8}

&
& SRCC & PLCC
& SRCC & PLCC
& SRCC & PLCC \\
\midrule

\multirow{5}{*}{FR-PCQA}
& PointSSIM~\cite{9106005}
& $0.738$ & $0.777$
& $0.220$ & $0.262$
& $0.932$ & $0.900$ \\

& PSNR-YUV~\cite{10.1117/12.2322741}
& $0.744$ & $0.732$
& $0.739$ & $0.738$
& $0.970$ & $0.978$ \\

& MSE-p2p~\cite{8296925}
& $0.761$ & $0.764$
& $0.625$ & $0.672$
& $0.940$ & $0.828$ \\

& MSE-p2plane~\cite{8296925}
& $0.636$ & $0.691$
& $0.629$ & $0.672$
& $0.939$ & $0.861$ \\

& PCQM~\cite{9123147}
& $0.735$ & $0.779$
& $0.785$ & $0.769$
& $0.967$ & $0.951$ \\

& FRSVR~\cite{Watanabe2025FRSVR}
& $0.601$ & $0.591$
& $0.610$ & $0.609$
& $0.970$ & $0.977$ \\

\midrule

\multirow{5}{*}{\shortstack{Supervised\\NR-PCQA}}
& GC-PCQA~\cite{10814096}
& $0.847$ & $0.866$
& $0.548$ & $0.576$
& $0.848$ & $0.918$ \\

& CLIP-PCQA~\cite{Liu2025CLIPPCQA}
& $0.966$ & $0.963$
& $0.353$ & $0.382$
& $0.633$ & $0.663$ \\

& GMS-3DQA~\cite{10.1145/3643817}
& $0.942$ & $0.926$
& $0.805$ & $0.846$
& $0.957$ & $0.975$ \\

& MM-PCQA~\cite{ijcai2023p195}
& $0.939$ & $0.948$
& $0.894$ & $0.894$
& $0.970$ & $0.967$ \\

& 3DTA~\cite{10542438}
& $0.926$ & $0.944$
& $0.931$ & $0.921$
& $0.894$ & $0.898$ \\

\midrule

\multirow{7}{*}{\shortstack{No PCQA-\\Specific Training}}
& NIQE~\cite{6353522}
& $0.497$ & $0.243$
& $0.125$ & $0.096$
& $0.460$ & $0.495$ \\

& IL-NIQE~\cite{7094273}
& $0.330$ & $0.364$
& $0.221$ & $0.224$
& $0.468$ & $0.575$ \\

& PIQE~\cite{7084843}
& $0.525$ & $0.523$
& $0.439$ & $0.428$
& $0.560$ & $0.587$ \\

& BRISQUE~\cite{6272356}
& $0.555$ & $0.576$
& $0.344$ & $0.167$
& $0.529$ & $0.617$ \\

& CLIP-IQA~\cite{Wang2023CLIPIQA}
& $0.052$ & $-0.048$
& $0.268$ & $0.269$
& $0.205$ & $0.199$ \\

& CLIP-IQA+~\cite{Wang2023CLIPIQA}
& $0.275$ & $0.255$
& $0.459$ & $0.445$
& $0.512$ & $0.531$ \\

& LIQE~\cite{Zhang_2023_CVPR}
& $0.502$ & $0.594$
& $0.678$ & $0.679$
& $0.371$ & $0.414$ \\


\rowcolor{gray!15}
& \textbf{Ours}
& \textbf{$0.721$} & \textbf{$0.739$}
& \textbf{$0.677$} & \textbf{$0.678$}
& \textbf{$0.712$} & \textbf{$0.718$} \\

\bottomrule
\end{tabular}
}
\end{table}

\subsection{Comparison with existing methods}
Table~\ref{tab:cmp} compares full-reference (FR) methods, supervised no-reference (NR) PCQA methods, and methods without PCQA-specific training. For the 2D quality metrics, we project the entire point cloud onto six orthographic views, apply the corresponding image-quality metric to each view, and average the scores as the estimated point cloud quality. 
We can observe from the table that the supervised NR-PCQA methods generally perform strongly because they learn directly from MOS labels, but their performance varies across databases with different contents and distortion processes. Without specific training, the proposed method outperforms most generic baselines on the three databases. This result suggests that reconstruction-induced discrepancies provide useful quality cues when geometric and projected-view responses are combined. The method does not match the strongest supervised or full-reference metrics in every setting, but it offers a practical reference-free alternative without PCQA-specific fine-tuning. Its larger performance gap on ICIP2020 suggests that the current reconstruction responses do not fully capture all codec-dependent quality variations.


\begin{table}[!ht]
\centering
\caption{Effect of complexity calibration on geometric, projected-view, and fused PCQA performance across three databases. ``w/ calib.'' and ``w/o calib.'' denote results with and without the calibration term, respectively. Bold values indicate the best performance within each branch.}
\label{tab:abl}

\resizebox{\columnwidth}{!}{
\begin{tabular}{c l cc cc cc}
\toprule

\multirow{2}{*}{Branch}
& \multirow{2}{*}{Variant}
& \multicolumn{2}{c}{SJTU-PCQA}
& \multicolumn{2}{c}{WPC2.0}
& \multicolumn{2}{c}{ICIP2020} \\

\cmidrule(lr){3-4}
\cmidrule(lr){5-6}
\cmidrule(lr){7-8}

&
& SRCC & PLCC
& SRCC & PLCC
& SRCC & PLCC \\

\midrule

\multirow{2}{*}{Fusion}
& w/ calib.
& $\mathbf{0.721}$ & $0.739$
& $\mathbf{0.677}$ & $\mathbf{0.678}$
& $0.712$ & $0.718$ \\

& w/o calib.
& $0.612$ & $0.644$
& $0.652$ & $0.653$
& $0.718$ & $0.714$ \\

\midrule

\multirow{2}{*}{Geometry}
& w/ calib.
& $0.712$ & $\mathbf{0.817}$
& $0.258$ & $0.273$
& $0.526$ & $0.618$ \\

& w/o calib.
& $0.652$ & $0.745$
& $0.328$ & $0.336$
& $0.604$ & $0.620$ \\

\midrule

\multirow{2}{*}{Projected view}
& w/ calib.
& $0.327$ & $0.414$
& $0.603$ & $0.607$
& $0.772$ & $\mathbf{0.812}$ \\

& w/o calib.
& $0.184$ & $0.371$
& $0.550$ & $0.555$
& $\mathbf{0.779}$ & $0.809$ \\

\bottomrule
\end{tabular}
}

\end{table}


\subsection{Ablation and calibration analysis}
Table~\ref{tab:abl} and Fig.~\ref{fig:content} examine the effects of complexity calibration and geometric--projected-view fusion. The geometric branch is more informative on SJTU-PCQA, which contains substantial geometric and mixed distortions. Its weaker performance on WPC2.0 may be related to the limited sensitivity of the geometric representation to attribute-quantization artifacts, which are explicitly included in this database. Whereas the projected-view branch captures these appearance changes. The two branches therefore provide complementary cues, and their fusion gives a more balanced prediction. Calibration improves both fused performance and MOS alignment in most cases.




\section{Conclusion}
\label{sec:conc}
We presented a reference-free point cloud quality assessment framework that uses successive surface reconstruction to calibrate reconstruction-induced discrepancies. The key idea is to use the change caused by stronger regularization as a content-dependent proxy for the component associated with intrinsic structural complexity. By combining calibrated geometric and projected-view cues, the proposed method achieves competitive performance without PCQA-specific training or fine-tuning.

Current geometric and appearance discrepancies are measured in frozen pretrained perceptual spaces. Although this simple instantiation provides an initial validation that reconstruction-induced discrepancies contain useful cues for PCQA, the calibration may not fully capture distortion types with reconstruction responses that differ substantially from those considered in the evaluated databases.  Future work will investigate point-cloud-specific measures of reconstruction-induced discrepancies, as well as content-adaptive reconstruction and calibration strategies. 

\bibliographystyle{IEEEbib}
\bibliography{refs}

\end{document}